\documentclass[a4paper,11pt]{article}
\pdfoutput=1
\usepackage{jheppub,bm,bbm}
\usepackage[T1]{fontenc}

\graphicspath{{./Figures/}}
\usepackage{algorithm}
\usepackage[noend]{algpseudocode}

\newcommand{\x}{\mathbf{x}}
\newcommand{\y}{\mathbf{y}}

\newcommand{\wh}[1]{\widehat{#1}}
\newcommand{\wt}[1]{\widetilde{#1}}

\newcommand{\rme}{\mathrm{e}}
\newcommand{\rmd}{\mathrm{d}}
\def\ba#1\ea{\begin{align}#1\end{align}}
\def\mkakko#1{\left(#1\right)}
\def\ckakko#1{\left\{#1\right\}}
\def\kkakko#1{\left[#1\right]}

\newcommand{\E}{\mathbb{E}}
\newcommand{\bphi}{\bm\phi}

\newcommand{\T}{\textsf{T}}

\newcommand{\1}{\mathbbm{1}}

\newcommand\Foo{{_1\text{F}_1}}
\newcommand\MVT{\text{MVT}}

\title{One-parameter family of acquisition functions for efficient global optimization}

\author{Takuya Kanazawa}
\affiliation{Research and Development Group, Hitachi, Ltd., Kokubunji, Tokyo 185-8601, Japan}
\emailAdd{takuya.kanazawa.cz@hitachi.com}

\abstract{Bayesian optimization (BO) with Gaussian processes is a powerful methodology to optimize an expensive black-box function with as few function evaluations as possible. The expected improvement (EI) and probability of improvement (PI) are among the most widely used schemes for BO. There is a plethora of other schemes that outperform EI and PI, but most of them are numerically far more expensive than EI and PI. In this work, we propose a new one-parameter family of acquisition functions for BO that unifies EI and PI. The proposed method is numerically inexpensive, is easy to implement, can be easily parallelized, and on benchmark tasks shows a performance superior to EI and GP-UCB. Its generalization to BO with Student-t processes is also presented.}

\begin{document} 
\maketitle
\flushbottom

\section{Introduction}

In diverse fields of science and engineering, one frequently faces the need to know the optimum of a black-box function that is expensive to evaluate. In materials science, in order to determine an optimal composition of alloys one has to repeat manual experiments that cost time and money. In machine learning model building, one has to tune a number of hyperparameters of a model but testing the performance of a model on big data via cross validation takes hours or even days. Thus, a framework is needed that provides a systematic means to minimize the number of queries needed to reach the optimal solution. Bayesian optimization (BO) \cite{Kushner1964,Mockus1978,Jones1998} is a powerful methodology to seek an optimum of a black-box function without knowledge of its analytical properties, such as its gradient. It has been utilized to solve problems in robotics and control tasks, drug design, automated machine learning, and mechanical design  \cite{Brochu2010tutorial,Shahriari2016,Frazier2018tutorial}. 

There are two basic ingredients in BO. The first is surrogate model building.  Given a few data points of an unknown function, one tries to estimate the global shape of the function and its degree of uncertainty by assuming certain smoothness properties of the function. In this regard the most common choice is to use Gaussian processes (GP)  \cite{RW_GP_book}, which have a number of favorable analytical properties and flexibility. The second ingredient is an acquisition function $\alpha(\x)$ that specifies where to query the next given previous observations. $\alpha(\x)$ quantitatively measures the utility of observing a point $\x$ next. There are many different options available for $\alpha$. The most widely used one is the expected improvement (EI) \cite{Mockus1978,Jones1998}, which is simple, is easy to implement, and empirically performs well on various optimization tasks. The theoretical properties of EI have been studied in \cite{vazquez2010,Bull2011}. A comprehensive review on the miscellaneous generalizations of EI can be found in \cite{Zhan2020}. Other options for $\alpha$ include the probability of improvement (PI) \cite{Kushner1964}, upper confidence bound (UCB) \cite{Srinivas2012}, entropy search \cite{Villemonteix2009,Hennig2012}, predictive entropy search \cite{HernandezLobato2014}, max-value entropy search \cite{Hoffman2015,Wang2017}, knowledge gradient \cite{Frazier2008}, GP-DC \cite{Kanazawa2021}, and GP-MGC \cite{kanazawa2021MGC}. However, except for PI and UCB, most of these approaches are computationally more expensive than EI.

In this paper, we propose a new one-parameter family of acquisition functions. It includes both PI and EI as special cases. It is known that PI and EI tend to favor exploitation over exploration, which often causes the algorithms to get stuck in a local optimum. Our generalized acquisition function circumvents this pitfall by pursuing a better balance between exploration and exploitation. We show on multiple benchmark tasks that after calibrating the parameter appropriately, our algorithm performs equally well and in some cases even outperforms baselines including PI, EI and GP-UCB. Our algorithm is numerically cheap and easy to implement, thereby providing a new attractive tool for efficient BO. We also generalize the proposed algorithm to BO using Student-t processes (TP) \cite{shah2013,shah2014,tracey2018} and compare its performance with GP.

This paper is organized as follows. 
In section~\ref{sc:defnew} the new acquisition function is defined and its theoretical properties including its relationship to preceding methods are illustrated. In section~\ref{sc:one} proposed method is tested on simple 1D functions, and its superior performance over the conventional EI is demonstrated. In section~\ref{sc:two} the proposed method is applied to the global optimization task of six benchmark functions that are highly multi-modal and difficult to optimize. In section~\ref{sc:tp} the proposed method is generalized from GP to TP and its relative performance is quantitatively examined.  In section~\ref{sc:conc} we conclude the paper. 

\section{New acquisition functions}\label{sc:defnew}

\begin{figure}[tbh]
	\centering
	\includegraphics[width=.47\textwidth]{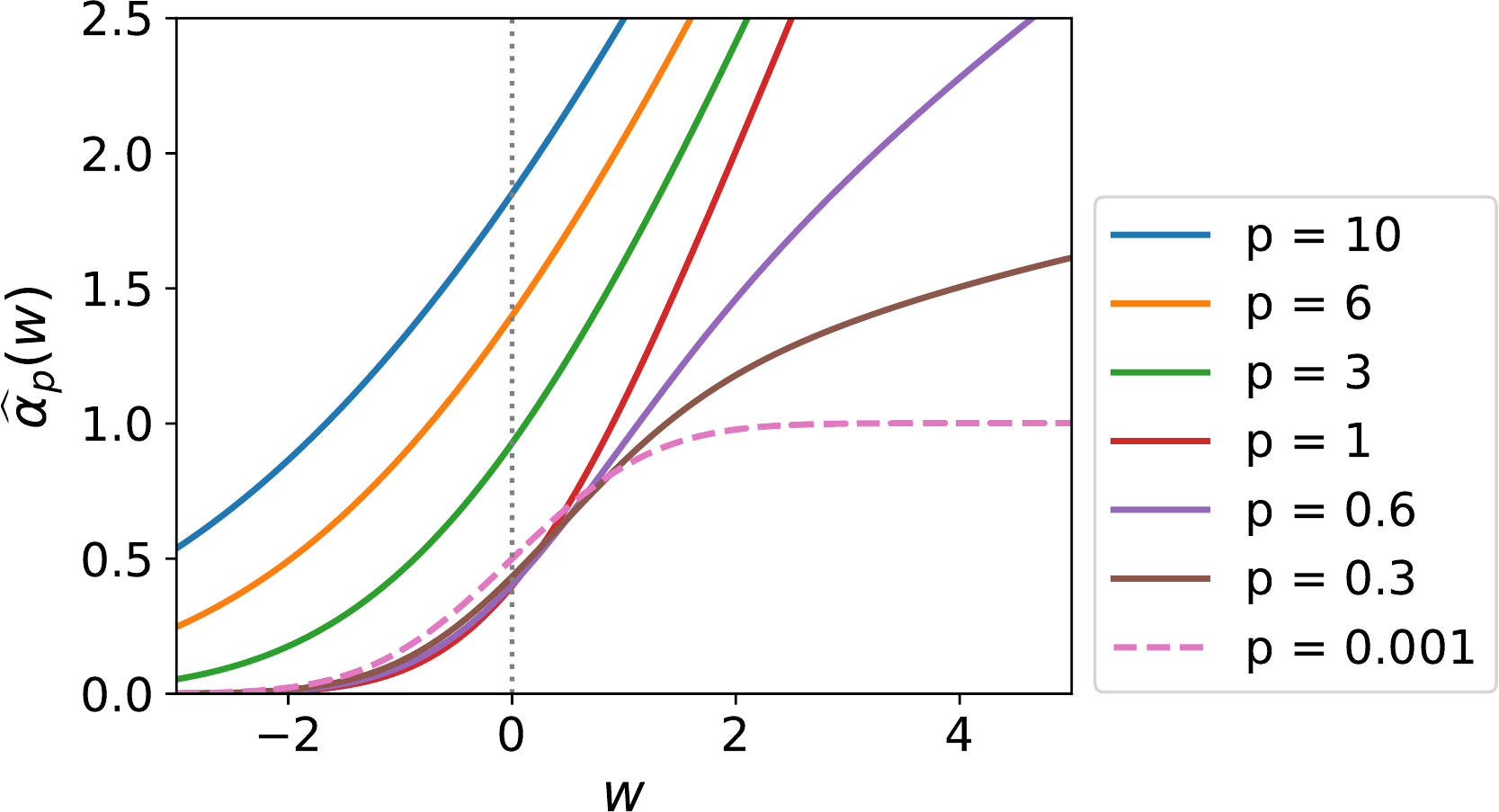}\quad 
	\includegraphics[width=.435\textwidth]{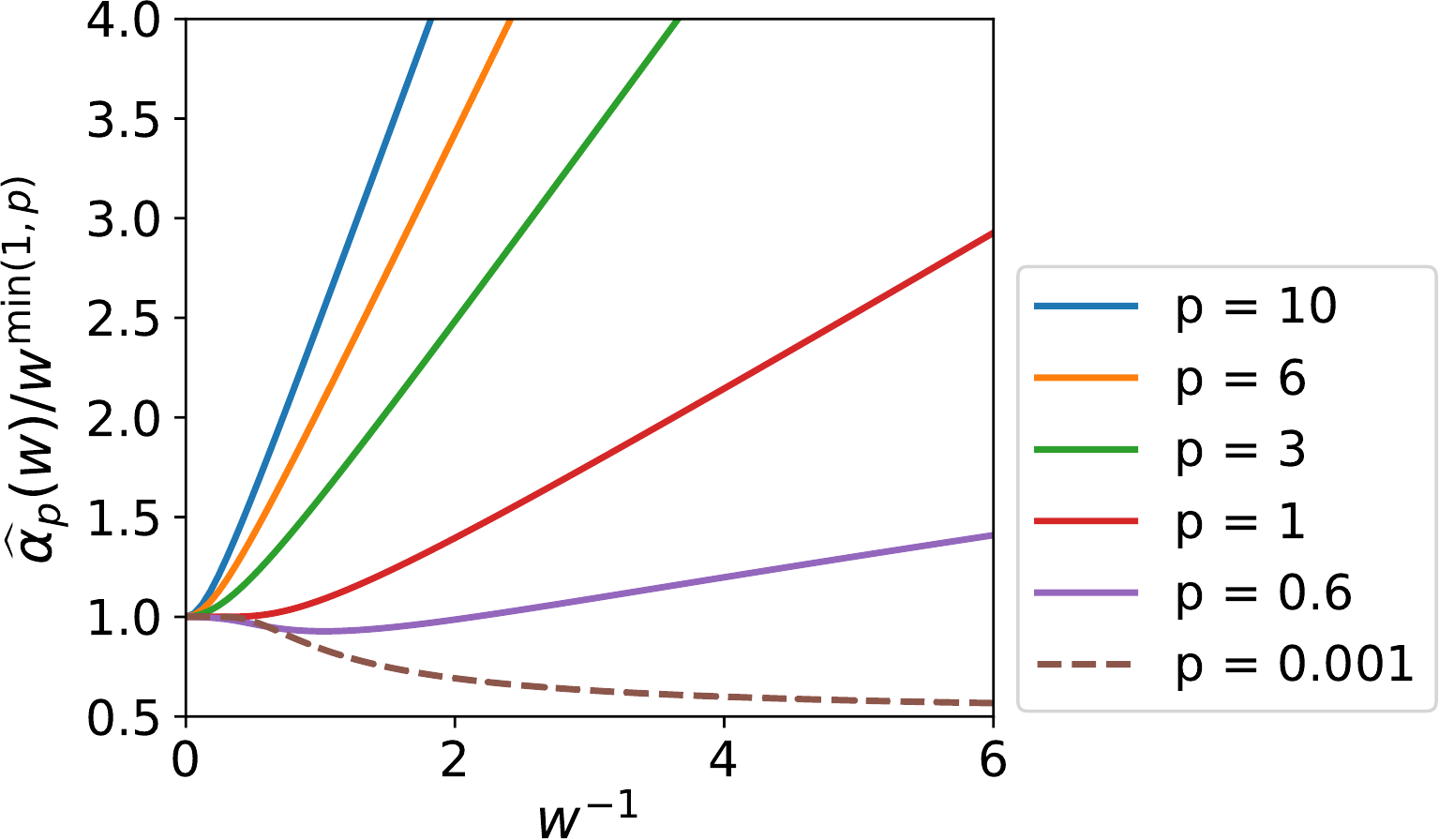}
	\caption{\label{fg:fds}Left:~The dimensionless acquisition function $\wh\alpha_p(w)$. Right:~$\alpha_p(\x)$ in units where $\mu(\x)-y_*=1$.}
\end{figure}

The acquisition function of EI is given by $\alpha_{\rm EI}(\x)=\E[(y-y_*)_+]$ where $(x)_+:=\max(x,0)$, $y_*$ is the maximum of the previously observed values, and the expectation value is taken with respect to the posterior distribution of the black-box function. Similarly, the acquisition function of PI is given by $\alpha_{\rm PI}(\x)=\mathrm{Prob}(y>y_*)=\E[\1(y>y_*)]$. Motivated by these standard definitions, we now introdue a novel one-parameter family of acquisition function labeled by $p>0$, defined as
\ba
	\alpha_p(\x) & \equiv \E\big[\ckakko{(y-y_*)_+}^p \big] 
	\\
	& = \frac{1}{\sqrt{2\pi \sigma(\x)^2}}\int_{y_*}^{\infty} \rmd y\; (y-y_*)^p ~
	\rme^{-\frac{[y-\mu(\x)]^2}{2\sigma(\x)^2}},
\ea
where $\mu(\x)$ and $\sigma(\x)$ are the predictive mean and predictive standard deviation, respectively. 
Introducing $z$ as $y=\mu(\x)+\sigma(\x)z$ we obtain
\ba
	\alpha_p(\x) & = \frac{1}{\sqrt{2\pi}}\int_{z_*}^{\infty} \rmd z\; 
	\kkakko{\mu(\x)+\sigma(\x)z-y_*}^p ~ \rme^{-z^2/2}
	\qquad \kkakko{z_* \equiv \frac{y_*-\mu(\x)}{\sigma(\x)}}
	\\
	& = \sigma(\x)^p \frac{1}{\sqrt{2\pi}}\int_{z_*}^{\infty}\rmd z\; 
	(z-z_*)^p~\rme^{-z^2/2}
	\\
	& = \sigma(\x)^p \frac{2^{\frac{p}{2}-1}}{\sqrt{\pi}}\kkakko{
		- \sqrt{2}z_*\Gamma\mkakko{\frac{p}{2}+1}
		\Foo\mkakko{\frac{1-p}{2},\frac{3}{2},-\frac{z_*^2}{2}}
		+ \Gamma\mkakko{\frac{p+1}{2}}
		\Foo\mkakko{-\frac{p}{2},\frac{1}{2},-\frac{z_*^2}{2}}
	}
\ea
where $\Foo$ is the confluent hypergeometric function. Note that $\alpha_p$ includes the acquisition functions of PI and EI as special cases
\ba
	\alpha_0(\x) & = \Phi(w)\big|_{w=-z_*}\,,
	\\
	\alpha_1(\x) & = \sigma(\x)\ckakko{
		\phi(w) + w \Phi(w)
	}\Big|_{w=-z_*},
\ea
where $\phi$ and $\Phi$ are the probability density and cumulative density of the standard normal distribution, respectively. For convenience, we also define the dimensionless function
\ba
	\wh\alpha_p(w) & = \mkakko{\frac{\alpha_p(\x)}{\sigma(\x)^p}}^{\min(p^{-1},1)}
	\\
	& = \ckakko{
	\frac{2^{\frac{p}{2}-1}}{\sqrt{\pi}}\kkakko{
		\sqrt{2}w\Gamma\mkakko{\frac{p}{2}+1}
		\Foo\mkakko{\frac{1-p}{2},\frac{3}{2},-\frac{w^2}{2}}
		+ \Gamma\mkakko{\frac{p+1}{2}}
		\Foo\mkakko{-\frac{p}{2},\frac{1}{2},-\frac{w^2}{2}}
	}}^{\min(p^{-1},1)}.
\ea
The power $\min(p^{-1},1)$ was chosen to avoid a singularity at $p=0$.  
$\wh\alpha_p(w)$ is plotted for various $p$ in Figure~\ref{fg:fds} (left).
As expected, it provides a smooth interpolation between PI ($p=0$) and EI ($p=1$). For all $p>0$, $\wh\alpha_p(w)$ is a monotonically increasing function. In units where $\sigma(\x)=1$, we have $w=-z^*=(\mu(\x)-y_*)/\sigma(\x)=\mu(\x)-y_*$, hence the monotonicity of $\wh\alpha_p(w)$ implies that points with larger gain $\mu(\x)-y_*$ are favored if the variance is the same. For $p\geq 1$, $\wh\alpha_p(w)$ asymptotes to the line $\wh\alpha_p(w)=w$ at large $w\gg 1$, while at $w<0$ we observe that $\wh\alpha_p$ with larger $p$ has a heavier tail than with small $p$. This implies that $\wh\alpha_p$ with larger $p$ tends to explore unobserved points actively, whereas $\wh\alpha_p$ with small $p$ exploits points with larger gain more agressively. To see this difference more clearly, let us look at the original acquisition function $\alpha_p$. If we employ units where the gain $\mu(\x)-y_*$ is unity so that $w=1/\sigma(\x)$, we have $\alpha_p(\x)^{\min(p^{-1},1)}=\sigma(\x)^{\min(1,p)}\wh\alpha_p(w)=\wh\alpha_p(w)/w^{\min(1,p)}$, which is plotted in Figure~\ref{fg:fds} (right). The horizontal axis $w^{-1}=\sigma(\x)$ represents the predictive uncertainty. One can see that for larger $p$ the function increases more rapidly, indicating enhanced preference for exploration over exploitation. In contrast, PI $(p=0)$ favors points with \emph{smaller} uncertainty, indicating a strong tendency to over-exploit. Interestingly, for $0<p<1$, the function is not monotonic: it first decreases but then starts increasing for larger $\sigma(\x)$. 

\begin{figure}[tbh]
	\centering
	\includegraphics[height=0.3\textwidth]{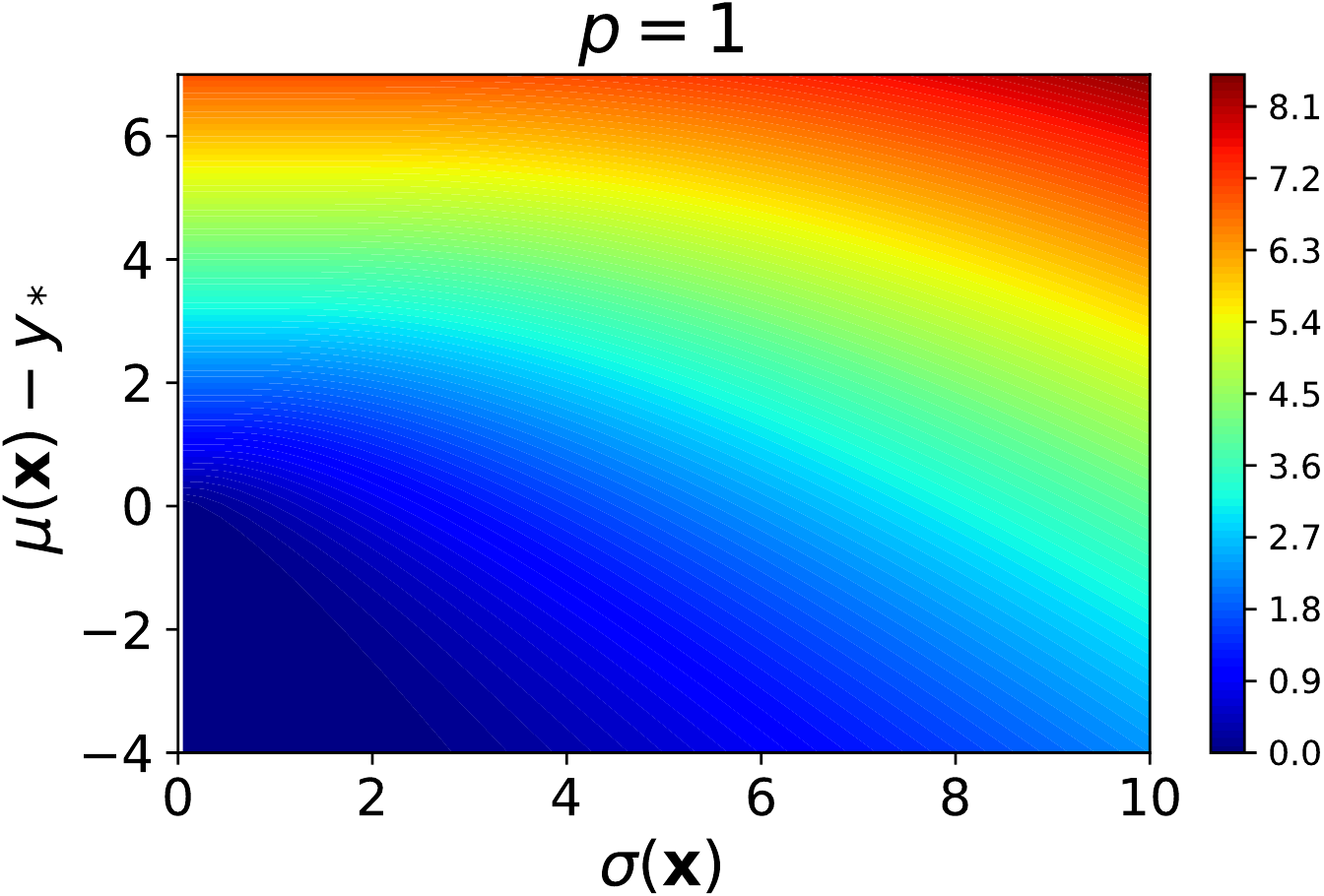}\quad 
	\includegraphics[height=0.3\textwidth]{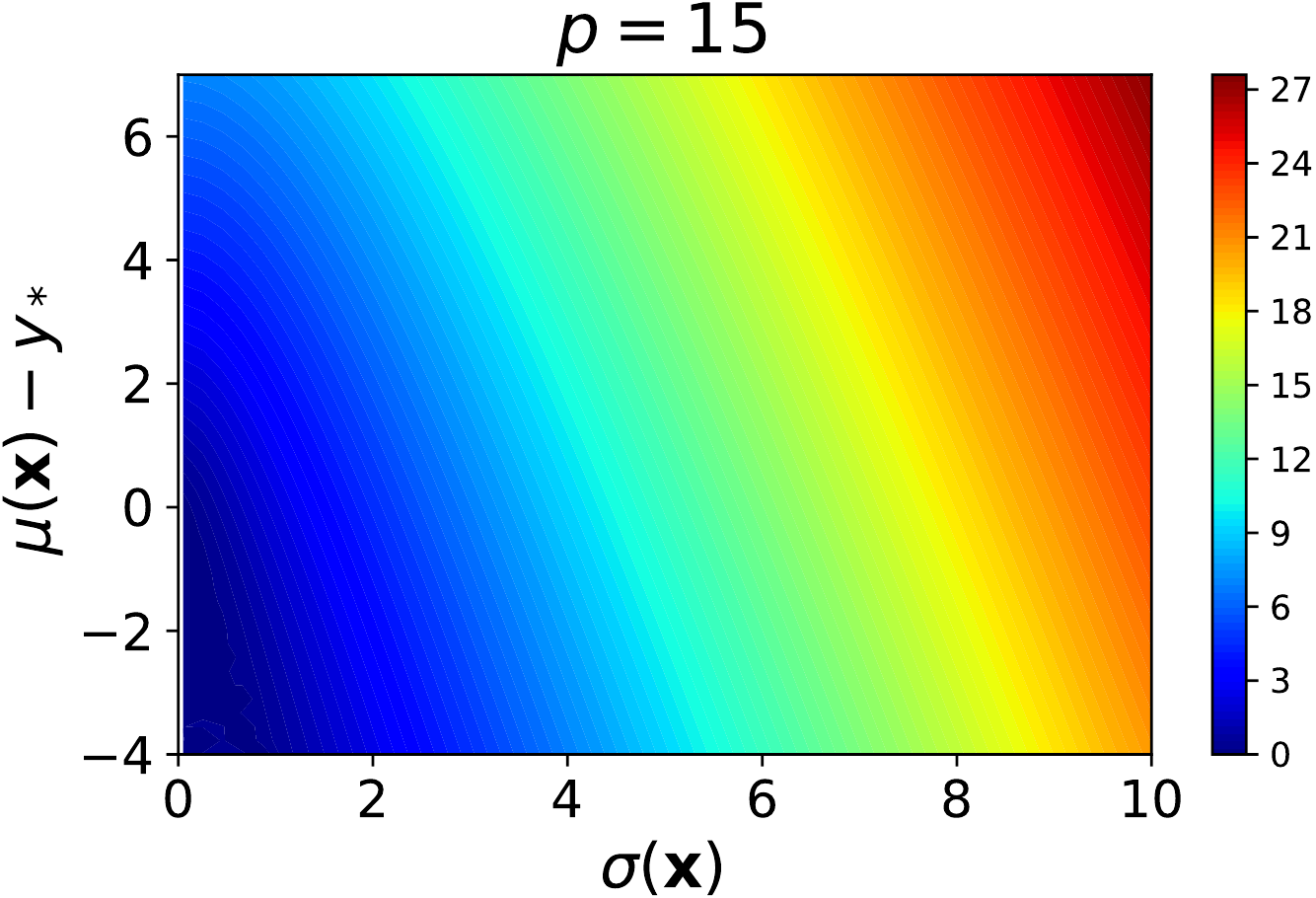}
	\caption{\label{fg:ppp}$\alpha_p(\x)^{1/p}$ as a function of $\sigma(\x)$ and $\mu(\x)-y_*$.}
\end{figure}

One can also view $\alpha_p$ as a function of two variables, the gain $\mu-y_*$ and the uncertainty $\sigma$. The plot for $p=1$ and $p=15$ are compared in Figure~\ref{fg:ppp}. 
From the direction of the gradient we can clearly see that $p=15$ seeks points with large uncertainty much more aggressively than $p=1$. Note that $\alpha_p(\x)$ for $p>1$ is an increasing function of both variables, which implies \cite{Greed2019} that the points chosen by $\alpha_p(\x)$ are always on the Pareto front. In this sense our approach mitigates the over-exploiting tendency of EI more nicely than the $\varepsilon$-greedy policy proposed in \cite{Bull2011} that chooses a point at random with probability $\varepsilon$ regardless of whether or not that point is Pareto optimal.

\section{Experiment on toy synthetic functions}
\label{sc:one}

\begin{figure}[tbh]
	\centering
	\includegraphics[width=0.4\textwidth]{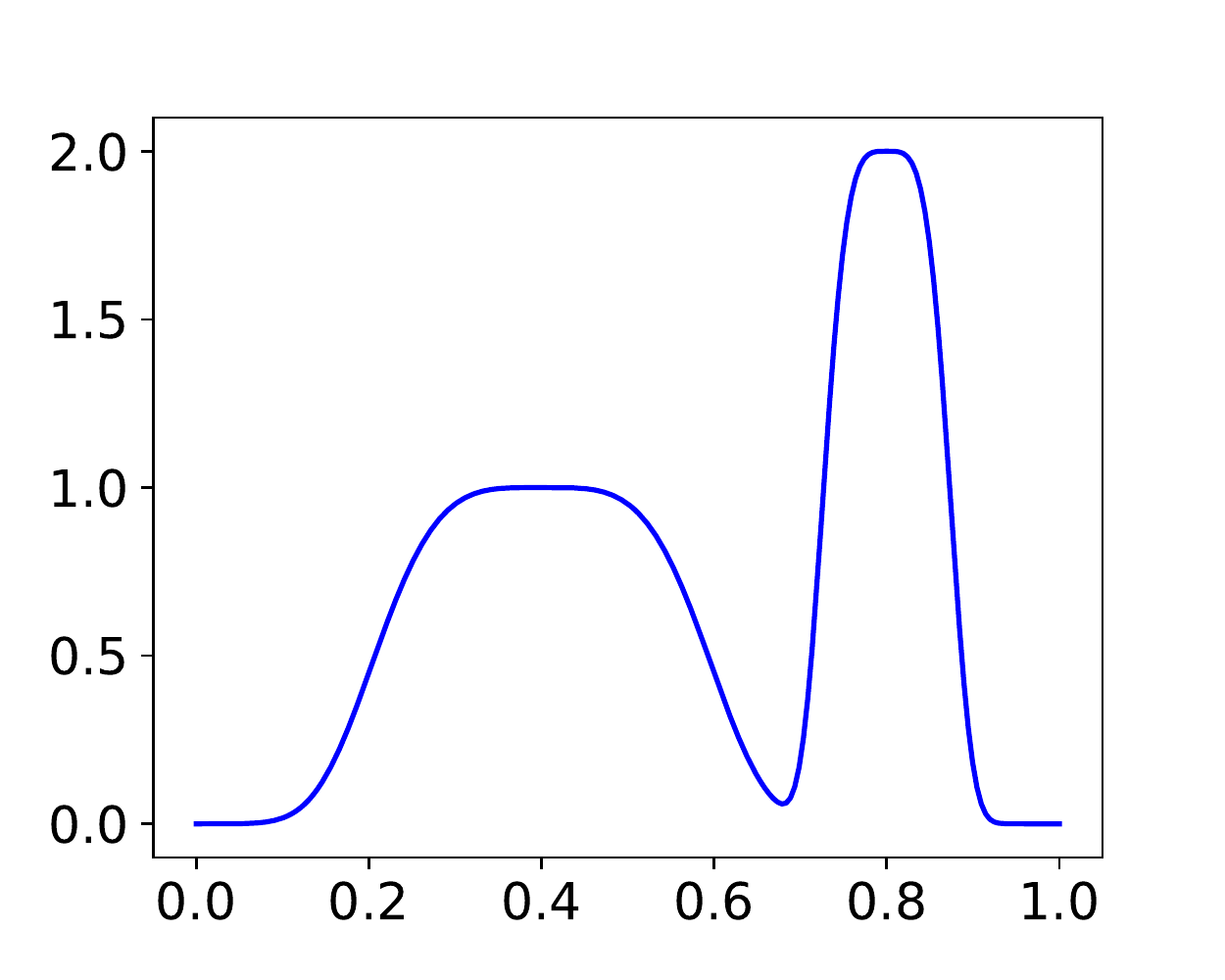}
	\quad
	\includegraphics[width=0.4\textwidth]{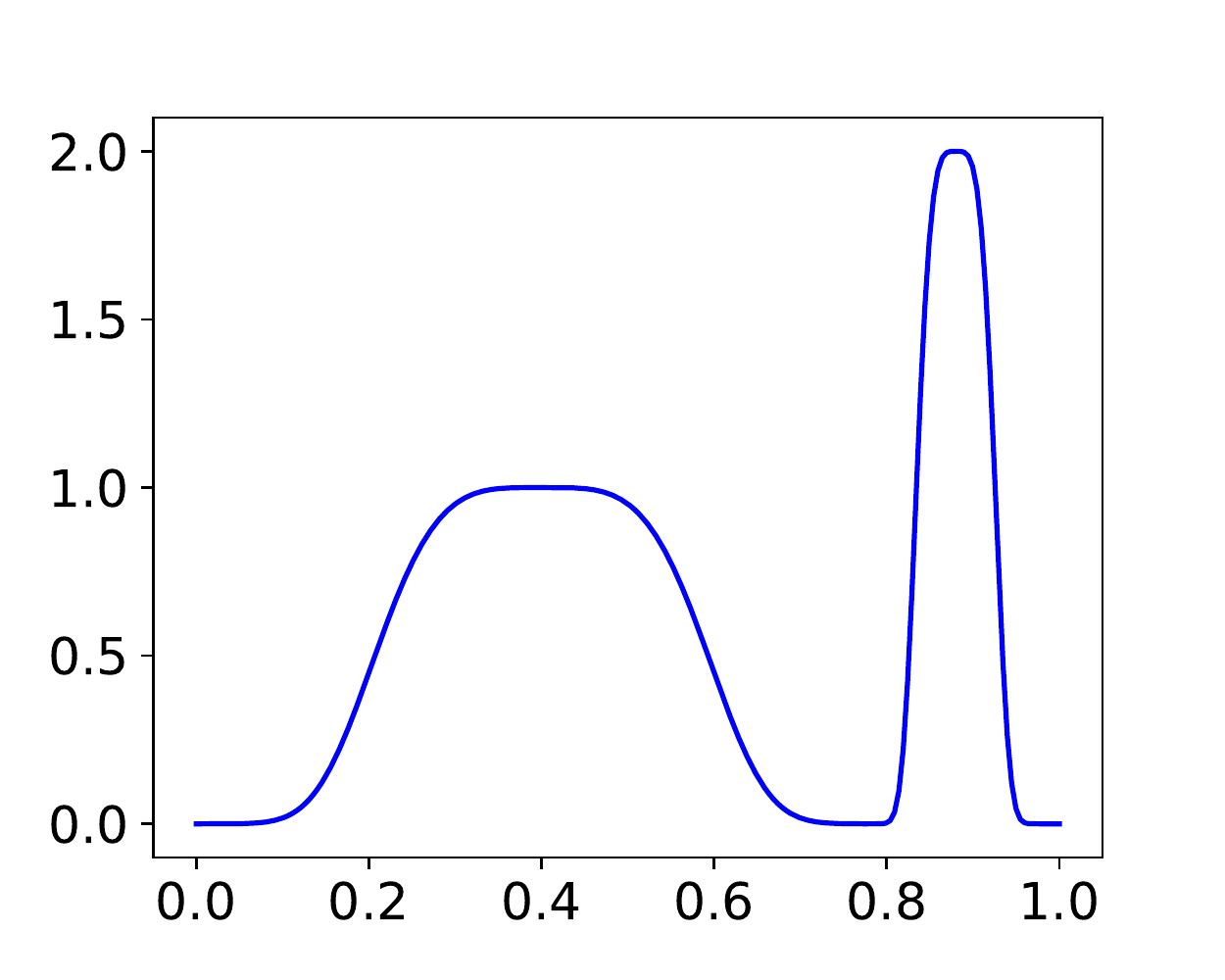}
	\put(-327,100){\Large $f_1$}
	\put(-140,100){\Large $f_2$}
	\caption{\label{fg:toy}Two toy functions on the interval $[0,1]$. Note that the higher peak is narrower for the right case, posing a more challenging optimization problem.}
\end{figure}

In this section we present numerical experiments on toy 1D functions on the unit interval. Define
\ba
	f_1(x) & = \exp[-500(x-0.4)^4] 
	+ 2\exp\bigg[-\mkakko{\frac{(x-0.8)}{0.08}}^4\bigg]\,,
	\\
	f_2(x) & = \exp[-500(x-0.4)^4] 
	+ 2\exp\bigg[-\mkakko{\frac{(x-0.88)}{0.05}}^4\bigg]\,.
\ea
Both have two peaks (see Figure~\ref{fg:toy}), with the narrower one being twice higher than the broader one.
The problem to be tackled by BO in this case is to avoid getting stuck in the local maximum and to reach the true global maximum. As one can see from Figure~\ref{fg:toy}, $f_1$ is a little easier to optimize than $f_2$. 

We conducted experiments as follows. At the beginning, the function values at two random points on $[0,1]$ are provided and the optimization process starts. In total $60$ new function evaluations are sequentially performed. At each step $T\leq 60$, the performance was measured by the best function value so far, i.e., $\displaystyle\max_t\{f(x_t)\}_{t=1:T}$. The score was averaged over 64 random initial conditions. The Mat\'{e}rn $5/2$ kernel was used, and the length scale and the overall magnitude of the kernel were adjusted after each function evaluation by maximizing the log marginal likelihood. 

\begin{figure}[tbh]
	\centering
	\includegraphics[width=.45\textwidth]{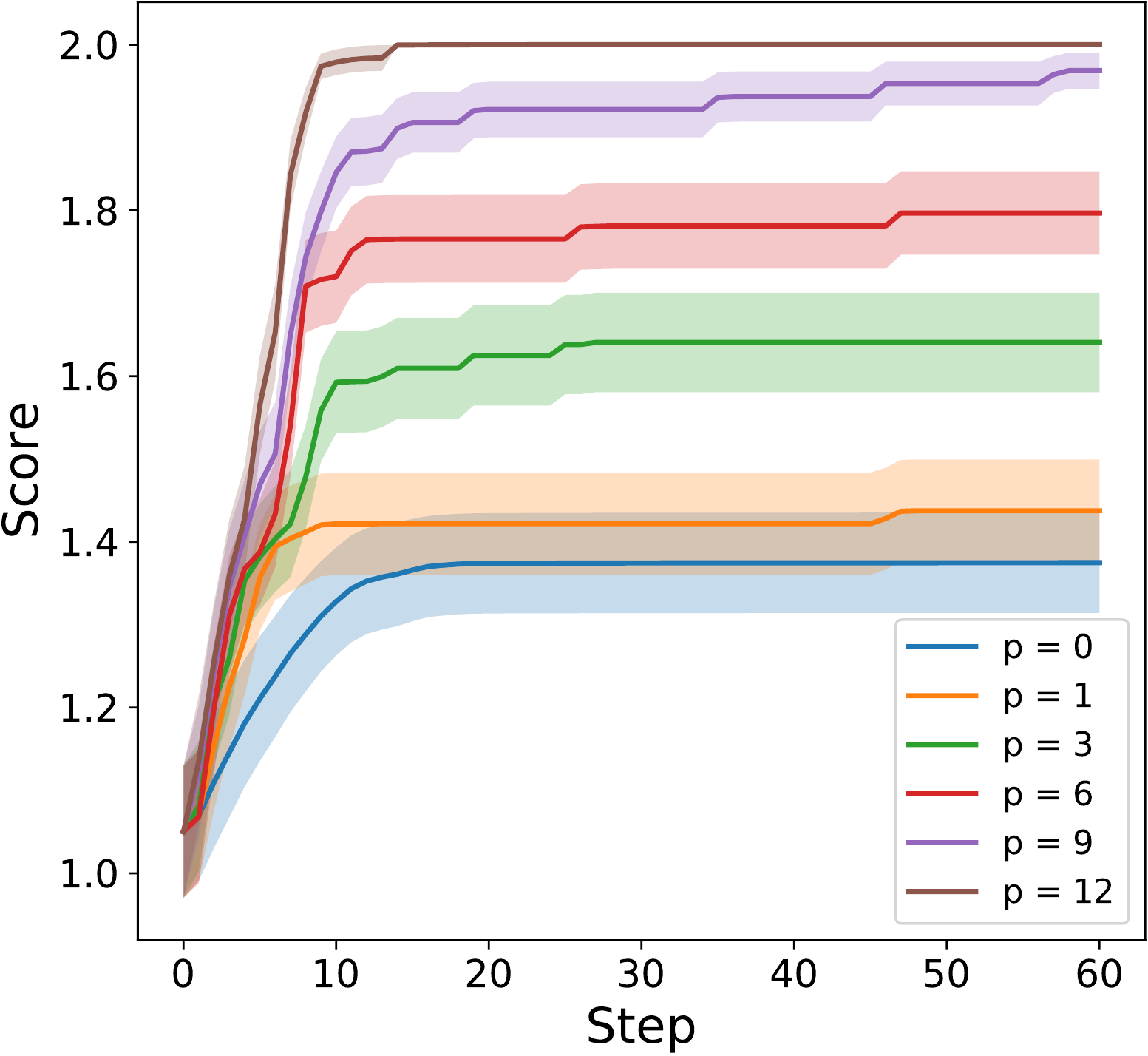}
	\quad 
	\includegraphics[width=.45\textwidth]{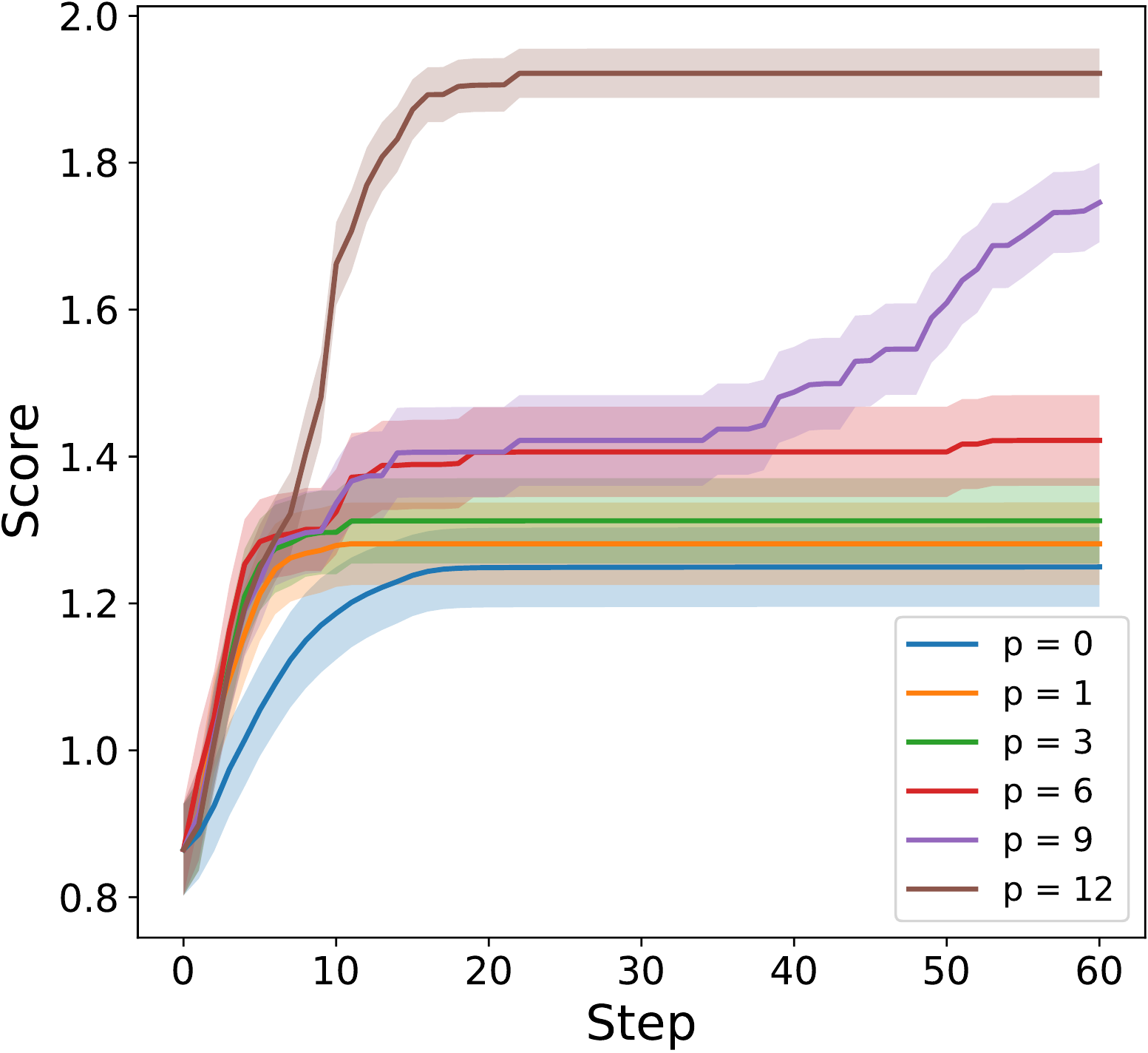}
	\caption{\label{fg:botoy}The result of numerical experiments using $\alpha_p$ on the function $f_1$ (left) and $f_2$ (right). Higher score is better. The shaded area represents one standard deviation.}
\end{figure}

The results are shown in Figure~\ref{fg:botoy}. Clearly, higher $p$ performs better while PI $(p=0)$ and EI $(p=1)$ fail to reach the true optimum with high probability.%
\footnote{This could be partly due to the fact that hyperparameters of the kernel function were updated sequentially during the course of optimization. The convergence proof of EI by Bull \cite{Bull2011} rests on the assumption that hyperparameters are fixed during optimization.} EI seems to stop exploration already at step 10. In contrast $p=12$ finds the true optimum of $f_1$ with 100\% probability. $f_2$ is more difficult to optimize than $f_1$, but $p=9$ and $12$ still succeed to find the true optimum of $f_2$ with high probability. This experiment illustrates that our new acquisition function has an enhanced optimization ability compared to the plain vanilla EI.

\section{Experiment on benchmark functions}
\label{sc:two}

In this section we test the utility of the proposed acquisition function on complex benchmark functions in higher dimensions. Below are the six functions considered here:
\begin{itemize}
	\setlength\itemsep{-3pt}
	\item Himmelblau function (2D)
	\item Eggholder function (2D)
	\item Hartmann function (3D)
	\item Ackley function (3D)
	\item Levy function (4D)
	\item Michalewicz function (4D)
\end{itemize}
These are all multimodal, exceptionally difficult to optimize functions. We refer to \cite{SB_website} for further details on the profile of each function. Although these functions are usually minimized, we took the negative of them and considered a maximization problem. All variables were rescaled so that each function was optimized in the unit hypercube $[0,1]^d$. 

We compared the performance of the following methods: 
(i)~Random search, (ii)~EI, (iii)~PI, (iv)~$\varepsilon$-EI \cite{Bull2011}, (v)~GP-UCB \cite{Srinivas2012}, and (vi)~BO with $\alpha_p(\x)$ for $p\in\{0.5, 2,3,4,6,8,10\}$. Here, $\varepsilon$-EI is the policy that adopts EI with probability $1-\varepsilon$ and chooses a random point with probability $\varepsilon$. We set $\varepsilon=0.1$. As for GP-UCB, we used Eq.~(5) in \citep{Brochu2010tutorial} with $\nu=1$ and $\delta=0.05$. We selected GP-UCB for this benchmarking because it has very strong theoretical guarantees for convergence to the global optimum \cite{Srinivas2012,freitas2012}. For each benchmark task, function values at 3 random points were initially provided, followed by 50 sequential function evaluations. The performance in the course of optimization was monitored via $\text{regret}_T :=\displaystyle\max_\x f(\x) - \max_{1\leq t\leq T}f(\x_t)$ ($T=1,2,\cdots,50$), which is the gap between the true optimum and the best value observed so far. Lower regret is better. The regret was averaged over 64 random initial conditions. In this section we assume noiseless observations. We used the Mat\'{e}rn $5/2$ kernel for all tasks following the recommendation of \cite{Snoek2012}, and updated the hyperparameters of the kernel after every observation via maximization of the log marginal likelihood. 

There are a variety of other methods that extend EI in miscellaneous ways, but most of them are numerically far more expensive than EI. For instance, \cite{Greed2019} requires multi-objective optimization; \cite{Berk2019} requires Thompson sampling from the posterior distribution; \cite{Qin2017} requires taking into account not only the predictive variance at respective candidate points but also the full covariance structure of the posterior distribution; and so on. Note also that the weighted expected improvement (WEI) of \cite{feng2015} is conceptually similar to the present work but WEI does not necessarily fulfill Pareto optimality as pointed out by \cite{Greed2019} and will not be considered here. 

\begin{figure}[htbp]
	\centering
	\includegraphics[height=.3\textheight]{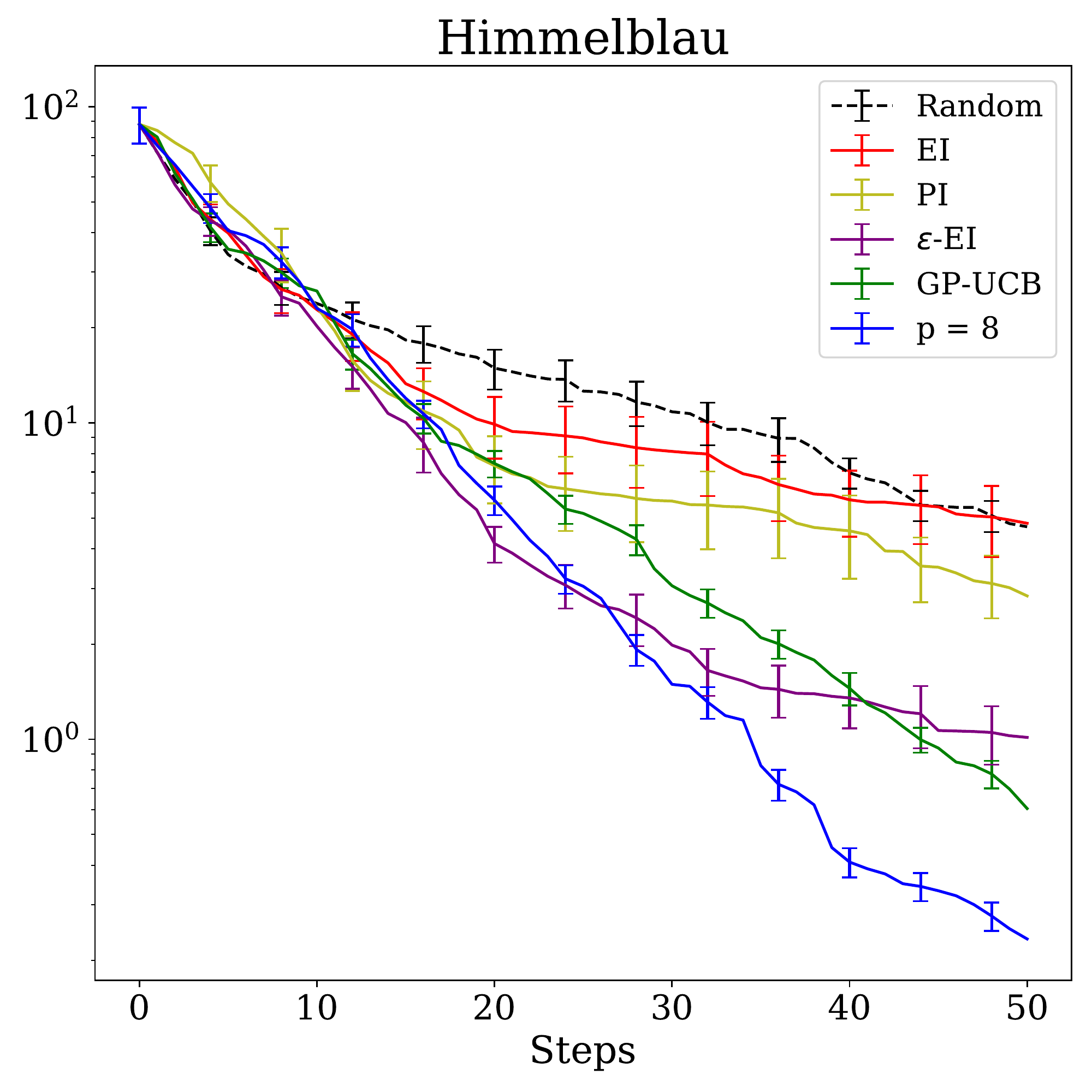} \quad 
	\includegraphics[height=.3\textheight]{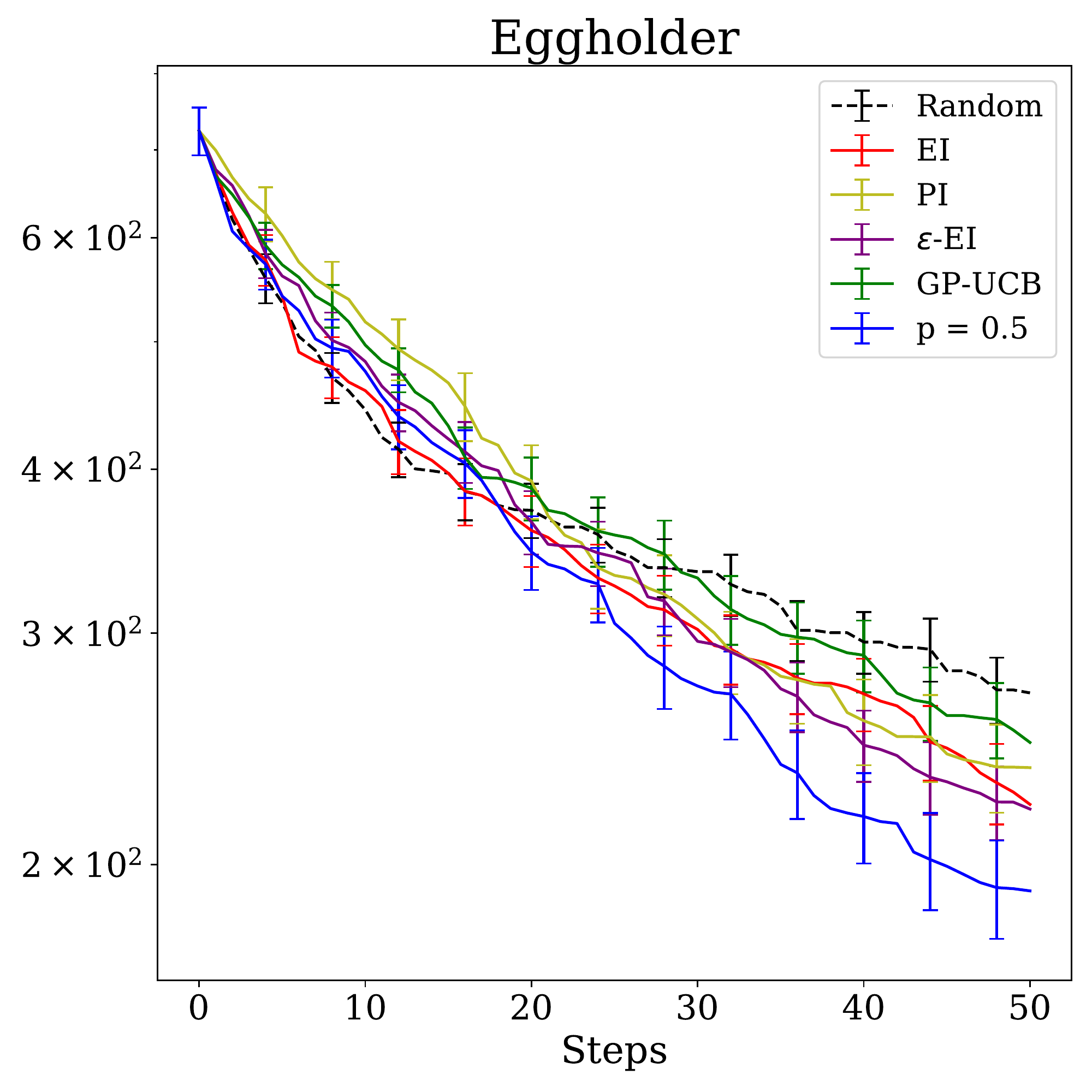}
	\vspace{\baselineskip}\\
	\includegraphics[height=.3\textheight]{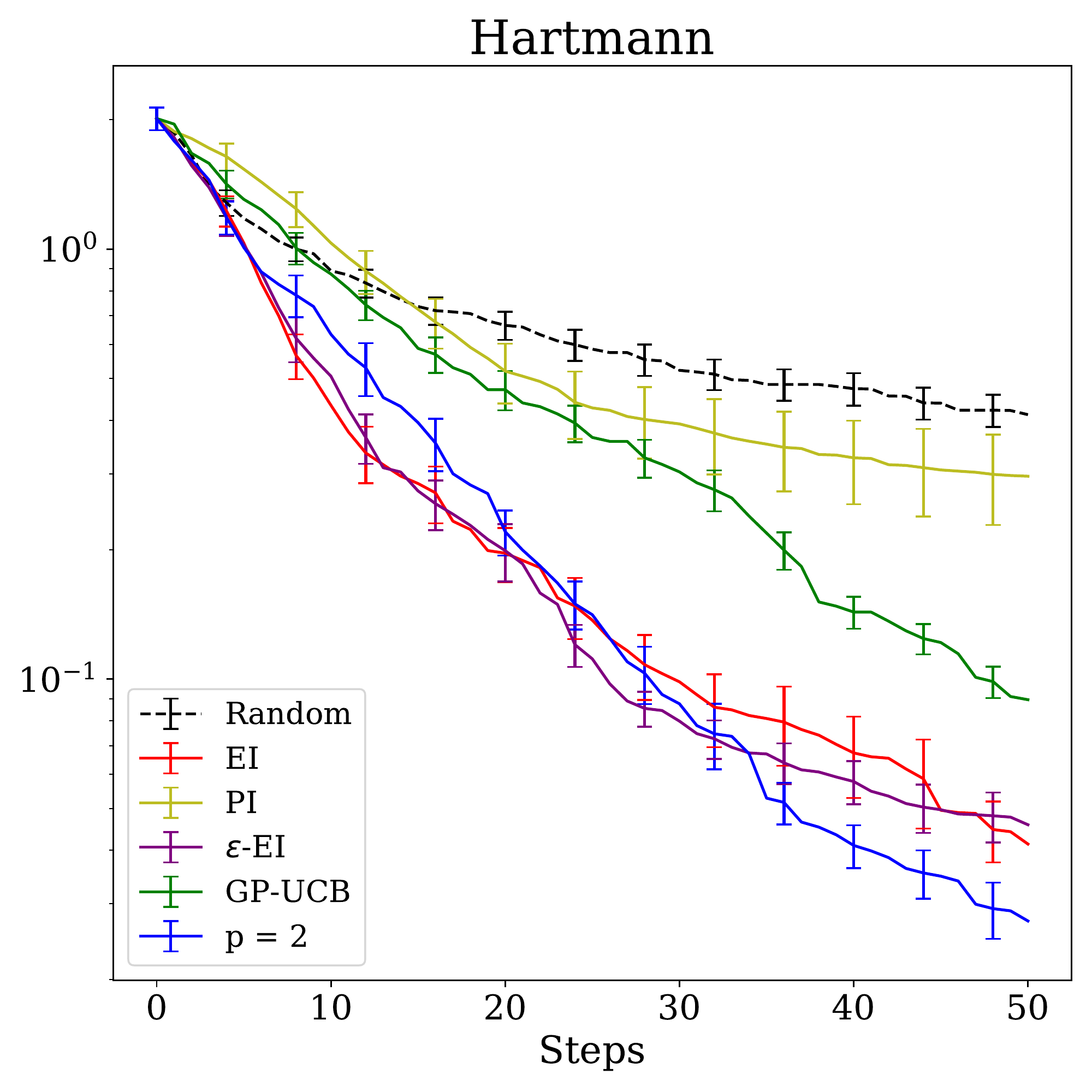} \quad 
	\includegraphics[height=.3\textheight]{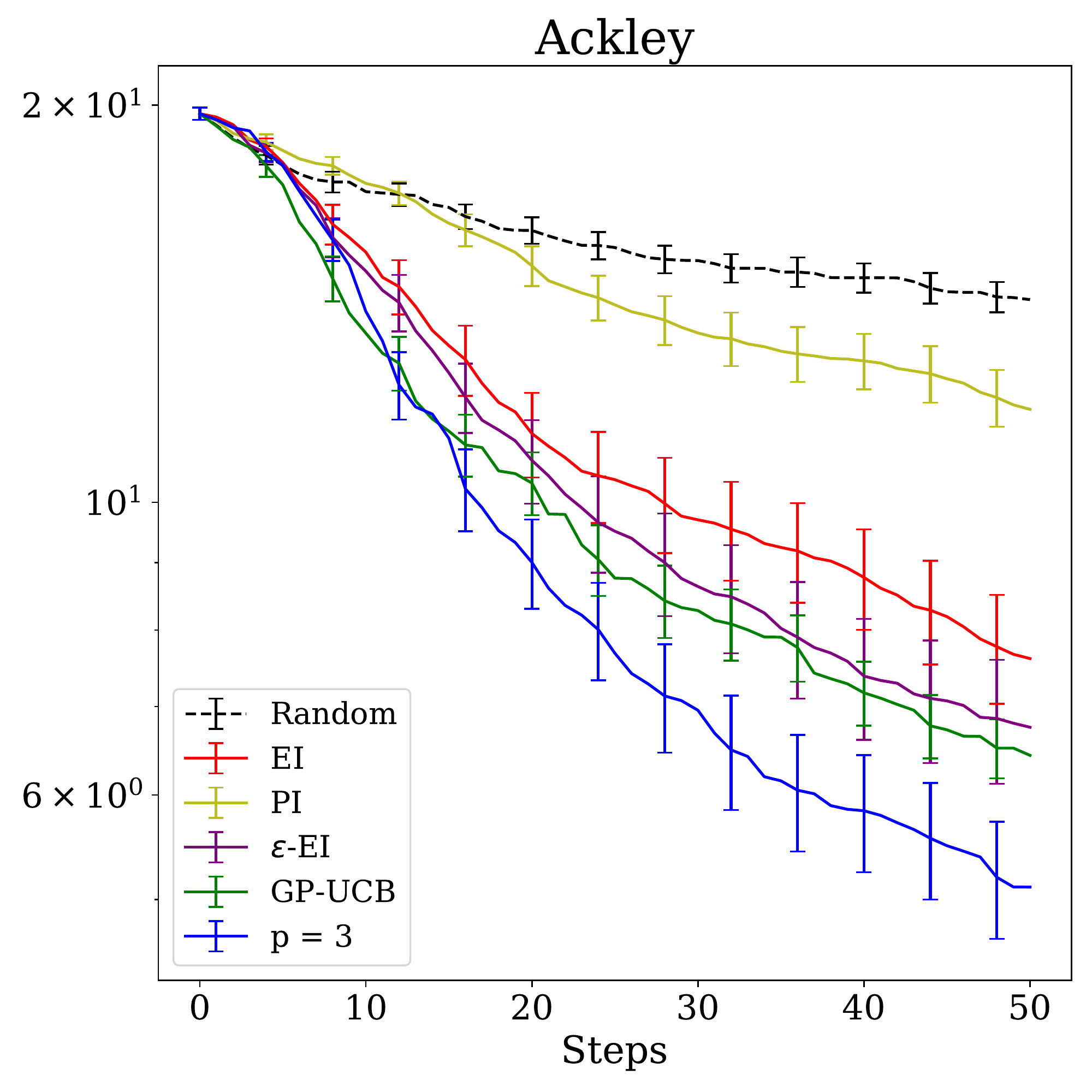}
	\vspace{\baselineskip}\\
	\includegraphics[height=.3\textheight]{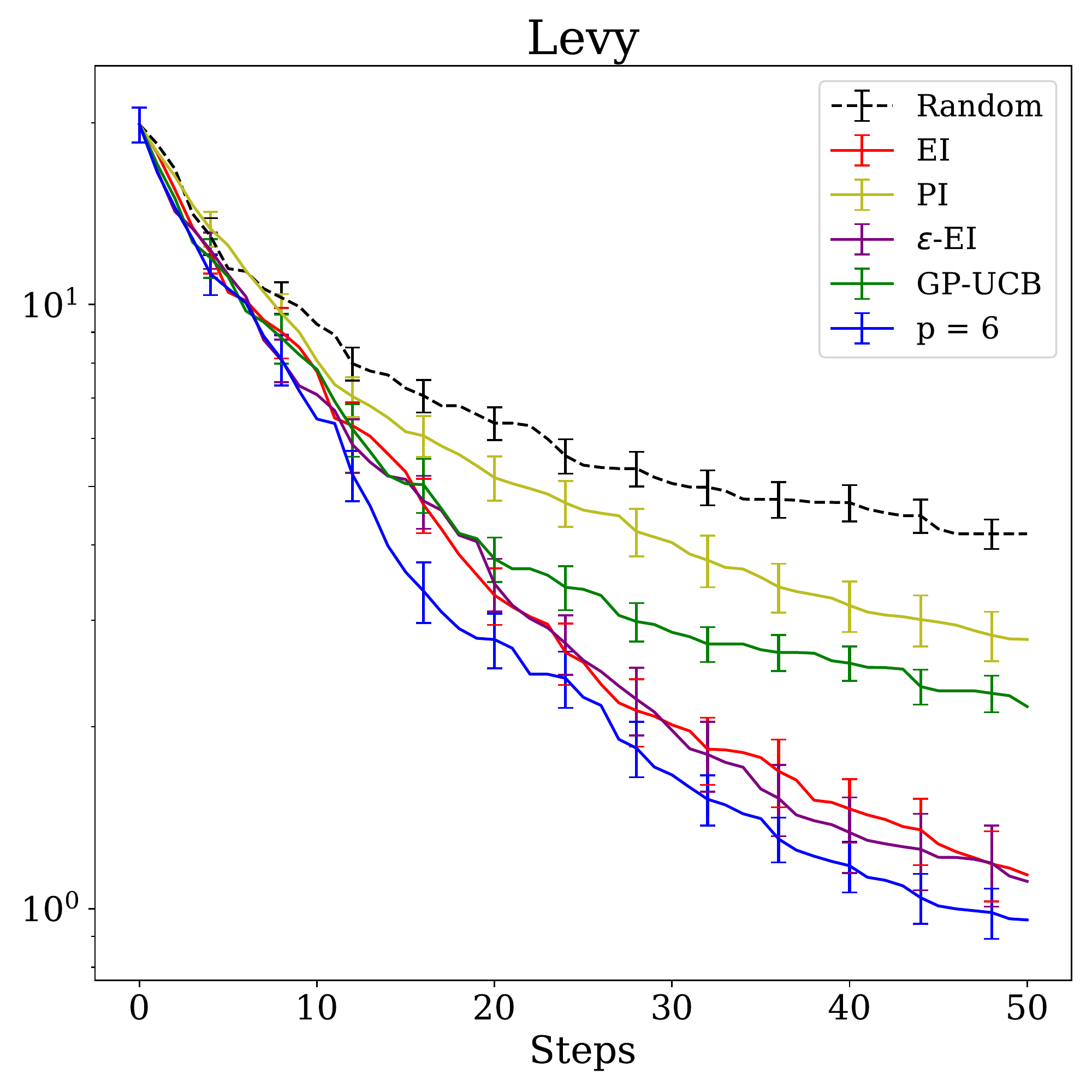} \quad 
	\includegraphics[height=.3\textheight]{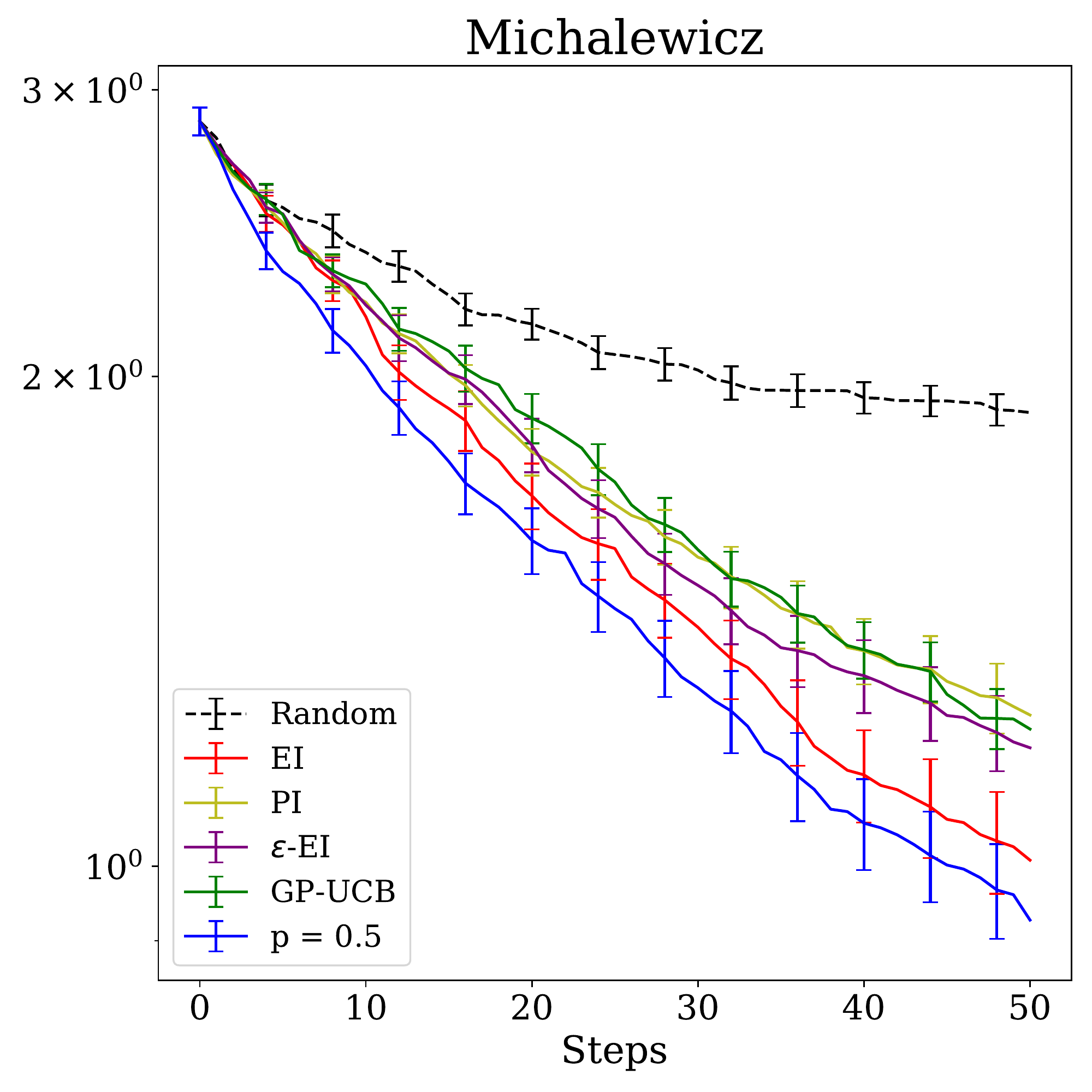}
	\caption{\label{fg:bench}Regret of BO with six policies as a function of the number of function evaluations on a log scale. The error bar represents one standard deviation of the mean.}
\end{figure}

The results of these numerical experiments are displayed in Figure~\ref{fg:bench}. To keep the visibility of the figures, we only plotted the $p$ that performed the best.  In all six tasks, the method using $\alpha_p$ with appropriate $p$ outperformed all the baselines. The improvement over EI is especially impressive for Himmelblau and Ackley functions. However, the calibration of $p$ is nontrivial. The best value of $p$ in Figure~\ref{fg:bench} ranges from $0.5$ to $8$ and is strongly task dependent. If one has prior knowledge over the function to be optimized, one can exploit it to tune $p$. Otherwise we recommend running multiple function evaluations in parallel using different values of $p$, which would be effective when resources for parallel computing are available. Another option is to vary $p$ during optimization. The optimal scheduling problem of $p$ seems to deserve a paper on its own and is deferred to future work.

\section{\boldmath Generalization to the Student-t processes}\label{sc:tp}

GP is widely used as a surrogate model for BO, but it has two drawbacks. First, the predictive distribution is always Gaussian and cannot account for outliers that come from heavy tails. Second, the predictive variance of GP is independent of functional values \cite{RW_GP_book}; it only depends on the locations of the observed points. Both of these  issues can be addressed by replacing GP with the Student-t processes (TP)  \cite{shah2013,shah2014,tracey2018,cantin2018,clare2020}.  
To introduce TP, we first define the multivariate Student-t distribution as \cite{shah2013,shah2014}
\ba
	p(\y) & = \frac{\Gamma(\frac{\nu+n}{2})}{\{(\nu-2)\pi\}^{\frac{n}{2}}\Gamma(\frac{\nu}{2})}
	|K|^{-1/2}\mkakko{
		1 + \frac{(\y-\bphi)^\T K^{-1}(\y-\bphi)}{\nu-2}
	}^{-(\nu+n)/2}
\ea
where $\y=(y_1,\cdots,y_n)^\T, \bphi=(\phi(\x_1),\cdots,\phi(\x_n))^\T$ and $K_{ij}=k(\x_i,\x_j)$. It holds that $\E[\y]=\bphi$ and $\text{cov}[\y]=K$. 
We write $\y \sim \MVT_n(\nu,\bphi,K)$. The parameter $\nu>2$ is called the degree of freedom. In the limit $\nu\to\infty$ $p(\y)$ converges to the Gaussian distribution. A point process is called TP if any finite collection of function values obey a joint multivariate Student-t distribution, $(f(\x_1),\cdots,f(\x_n))\sim\MVT_n(\nu,\bphi,K)$. 

It was empirically shown \cite{shah2013,shah2014,tracey2018} that BO utilizing TP in place of GP dramatically improves performance. While the authors of \cite{shah2013,shah2014,tracey2018} focused on EI as the infill criterion, TP can in principle be combined with any other criterion. Therefore, the following question is important to consider: can the optimization performance based on $\alpha_p$ be improved by using TP?

To derive the acquisition function for TP, we shall start from the fact \cite{shah2013,shah2014} that after $n$ observations $\y_n=(y_1,\cdots,y_n)^\T$ the predictive distribution at a new point $\x$ is given by
\ba
	y|\y_n \sim \MVT_1\mkakko{\nu+n, \wt{k}^\T K^{-1}(\y_n-\bphi_n)+\phi, 
	\frac{\nu+(\y_n-\bphi_n)^\T K^{-1}(\y_n-\bphi_n)-2}{\nu+n-2}(C-\wt{k}^\T K^{-1}\wt{k})}
\ea
where $\wt{k}=(k(\x,\x_1),\cdots,k(\x,\x_n))^\T$ and $C=k(\x,\x)$. Namely,
\ba
	p(y|\y_n) & = \frac{1}{\sqrt{V(\x)}}q_{\nu+n}\mkakko{\frac{y-\mu(\x)}{\sqrt{V(\x)}}}
\ea
where
\ba
	q_m(z) & \equiv \frac{\Gamma(\frac{m+1}{2})}{\sqrt{\pi}\Gamma(\frac{m}{2})}
	(1+z^2)^{-(m+1)/2},
	\\
	\mu(\x) & \equiv  \wt{k}^\T K^{-1}(\y_n-\bphi_n)+\phi\,,
	\\
	V(\x) & \equiv [\nu+(\y_n-\bphi_n)^\T K^{-1}(\y_n-\bphi_n)-2](C-\wt{k}^\T K^{-1}\wt{k})\,.
\ea
Then it follows from simple algebra that
\ba
	\alpha_p(\x) & \equiv \E\big[\ckakko{(y-y_*)_+}^p \big] 
	\\
	& = \int_{y_*}^{\infty}\rmd y\; (y-y_*)^p p(y|\y_n)
	\\
	& = \int_{y_*}^{\infty}\rmd y\; (y-y_*)^p \frac{1}{\sqrt{V(\x)}}q_{\nu+n}\mkakko{\frac{y-\mu(\x)}{\sqrt{V(\x)}}}
	\\
	& = V(\x)^{p/2} \int_{z_*}^{\infty}\rmd z\; (z-z_*)^p q_{\nu+n}(z) \qquad 
	\kkakko{z_* \equiv \frac{y_* - \mu(\x)}{\sqrt{V(\x)}}}. 
\ea
For convergence of the integral we need $p<\nu+n$, so we use $p\leftarrow \min(p,\lfloor \nu+n \rfloor)$ in place of $p$. The aforementioned integral can be easily evaluated using standard numerical routines.

To evaluate the performance of $\alpha_p$ combined with TP, we conducted numerical experiments using the same six benchmark functions as in the last section. We placed a log-normal prior on $\nu-2$ following \cite{shah2013} and updated it after every function evaluation via the maximum log likelihood method. The benchmarking was repeated for 64 random initial conditions. To compare TP with GP, we computed the index
\ba
	\frac{1}{50}\sum_{t=1}^{50}\log
	\mkakko{\frac{\text{regret}_t^{\rm (TP)}}{\text{regret}_t^{\rm (GP)}}}
	\label{eq:fdsfp}
\ea
for each random initial condition and for each $p$, where 50 is the number of sequential function evaluations that follow after random initialization. The negative value of the index indicates that TP performs better than GP, and vice versa. 

\begin{figure}[tbp]
	\centering
	\includegraphics[height=.23\textheight]{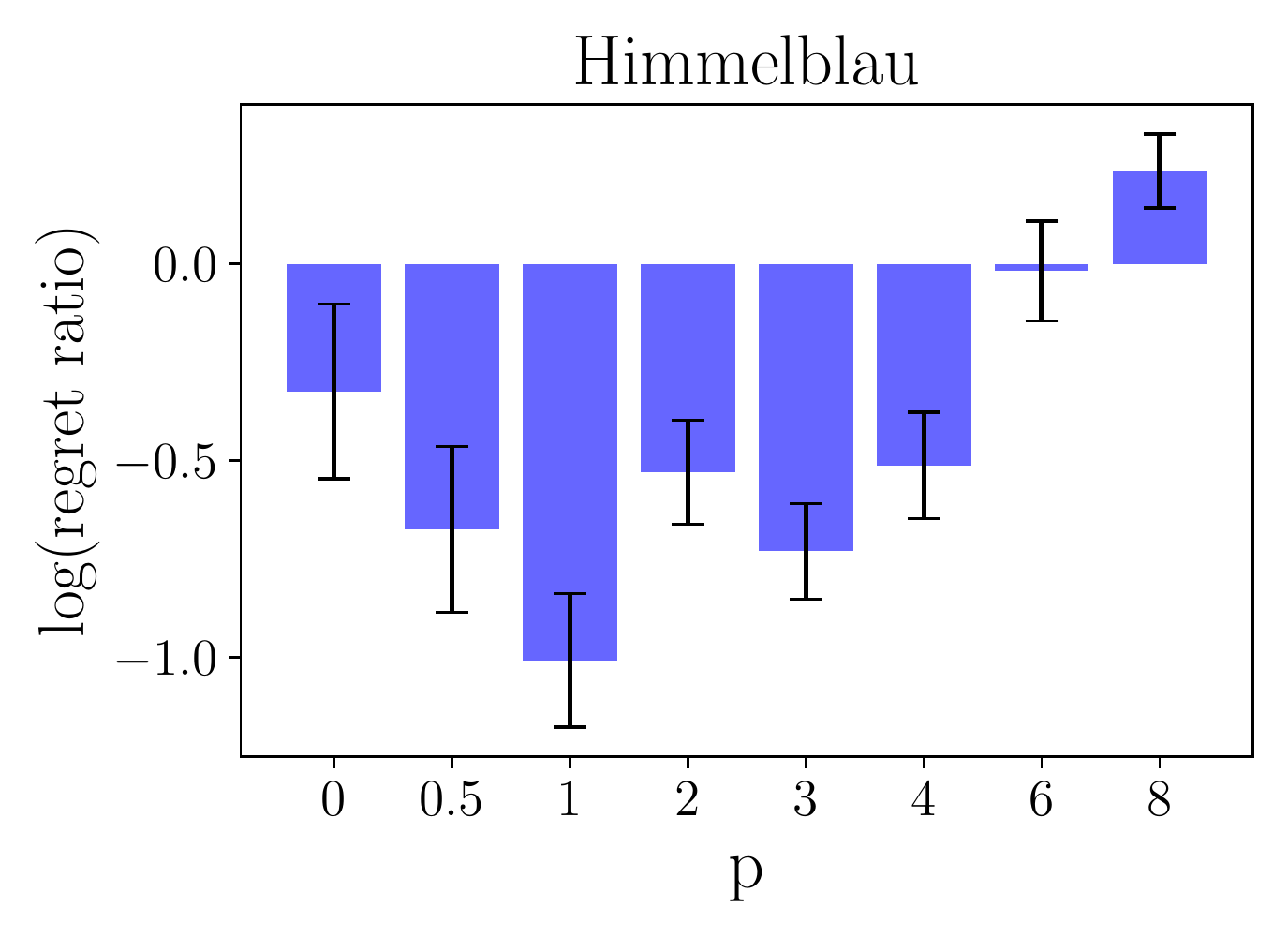}
	\includegraphics[height=.23\textheight]{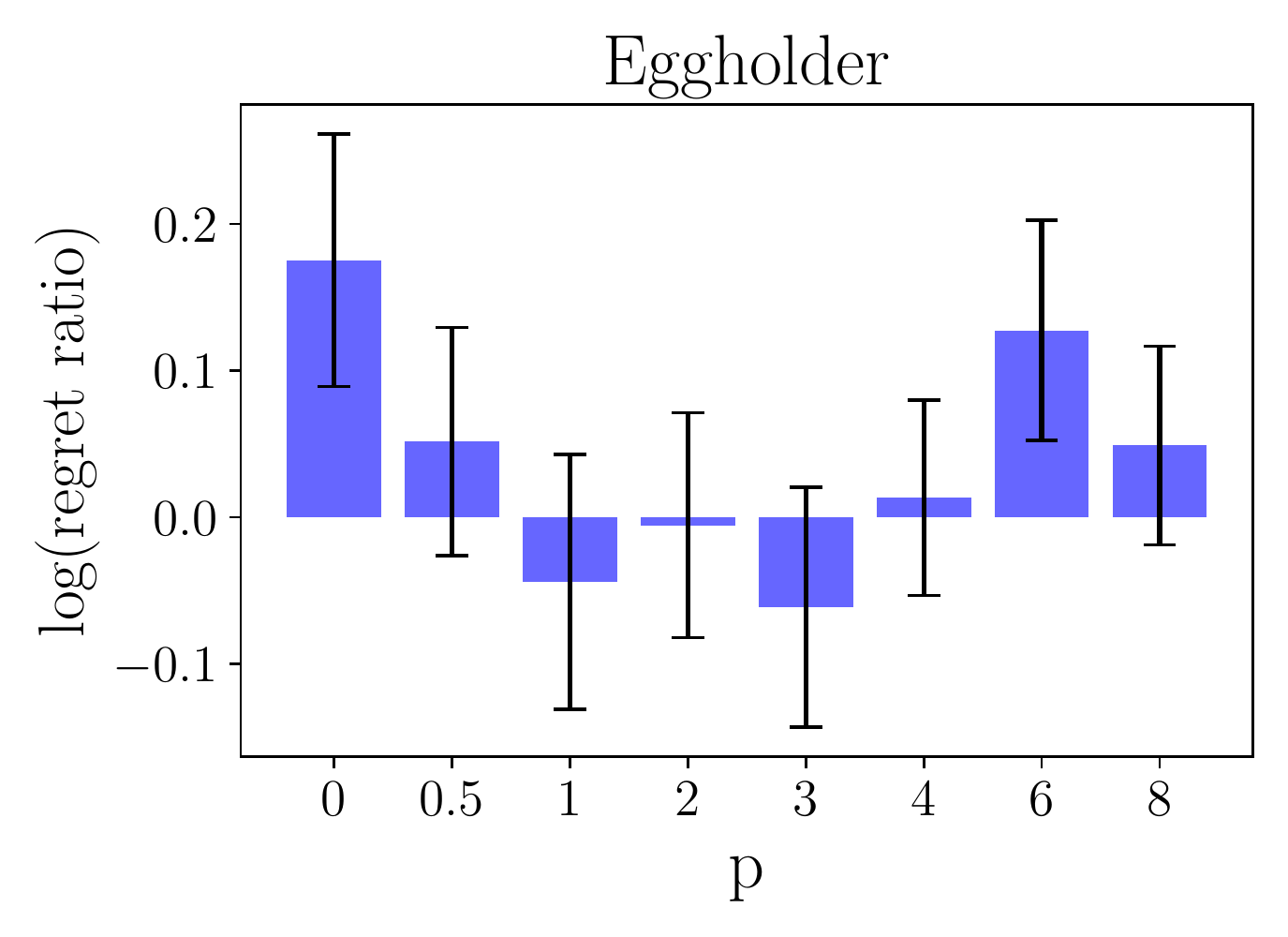}
	\vspace{3mm}\\
	\includegraphics[height=.23\textheight]{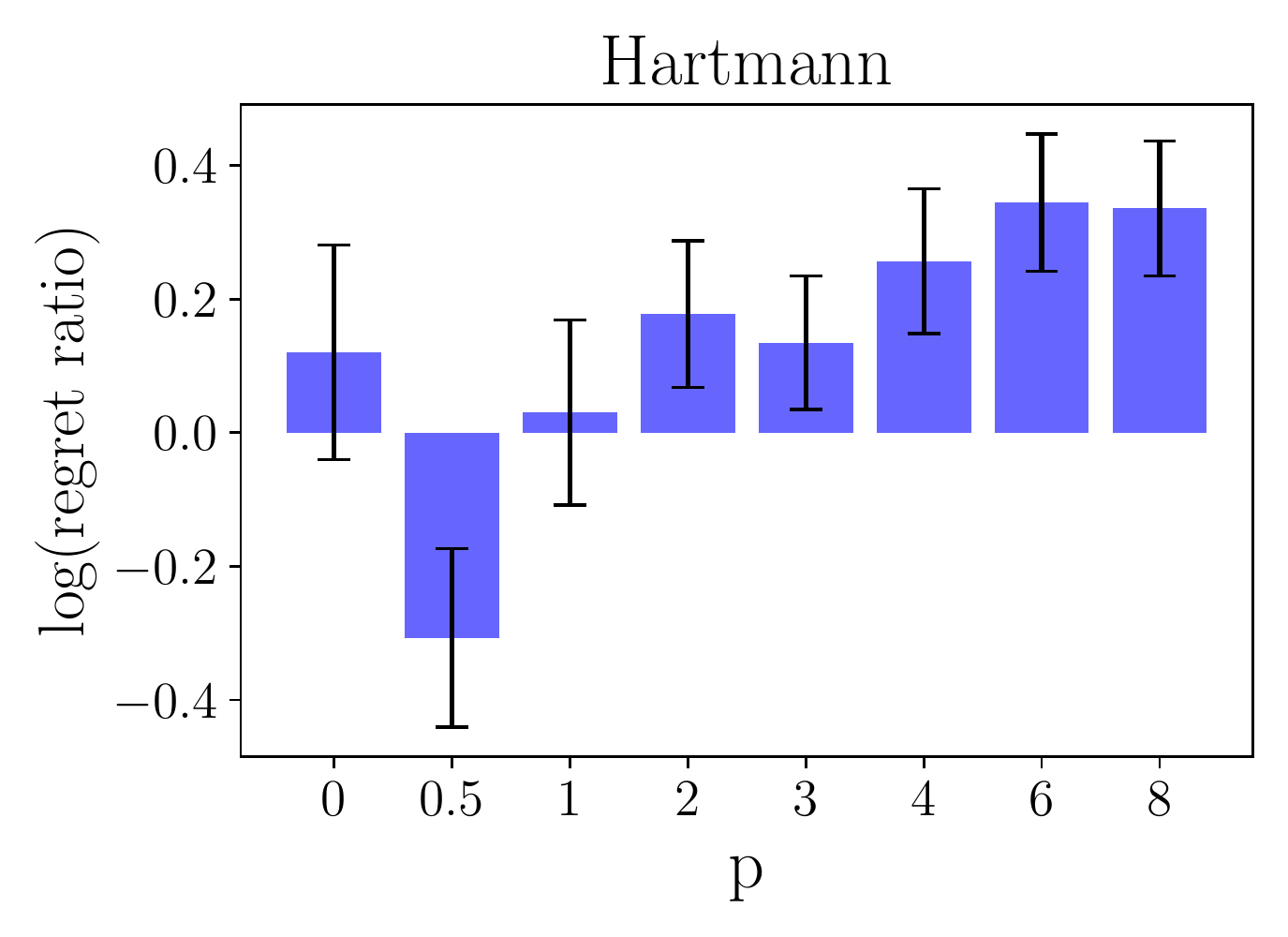}
	\includegraphics[height=.23\textheight]{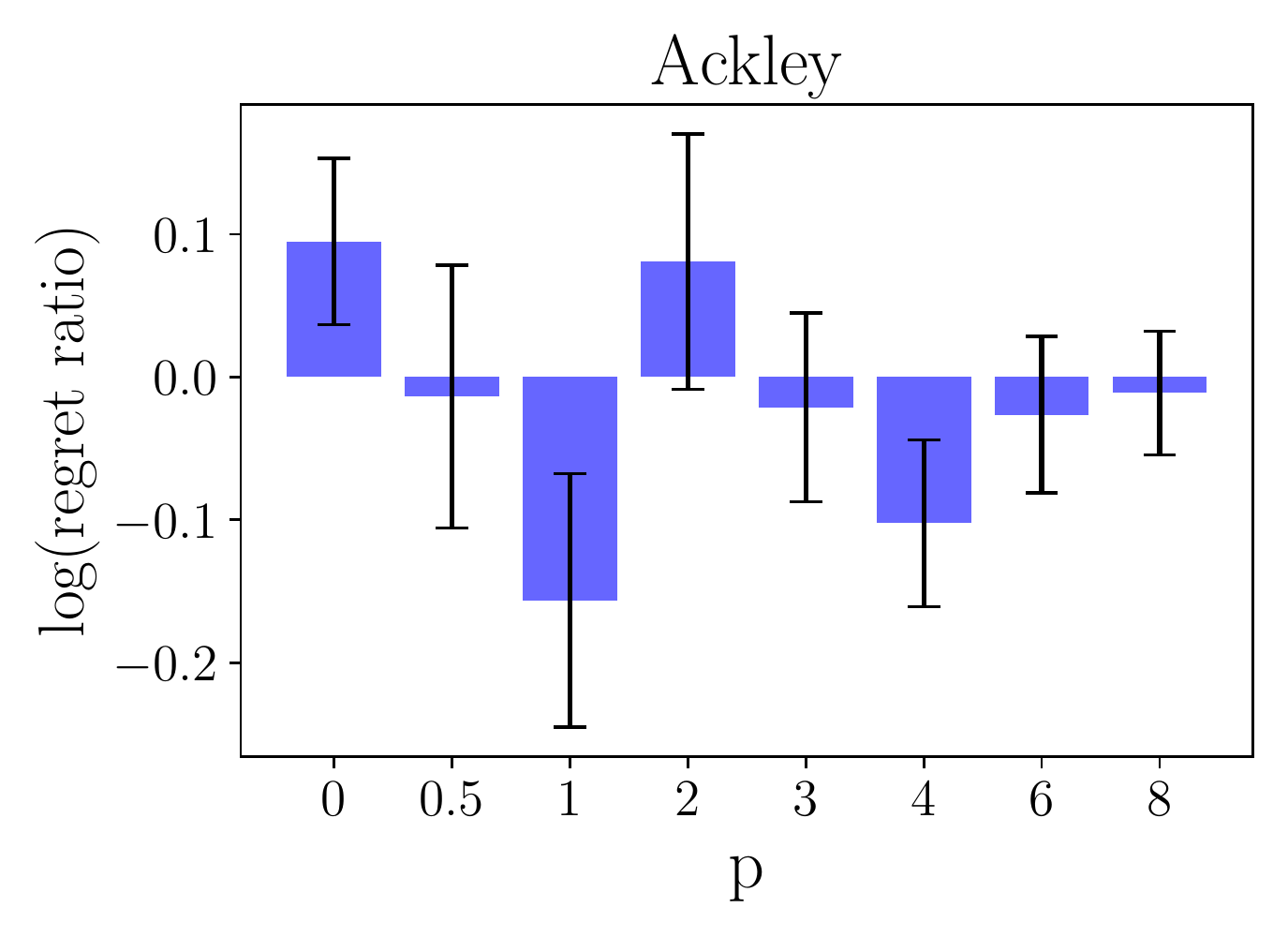}
	\vspace{3mm}\\
	\includegraphics[height=.23\textheight]{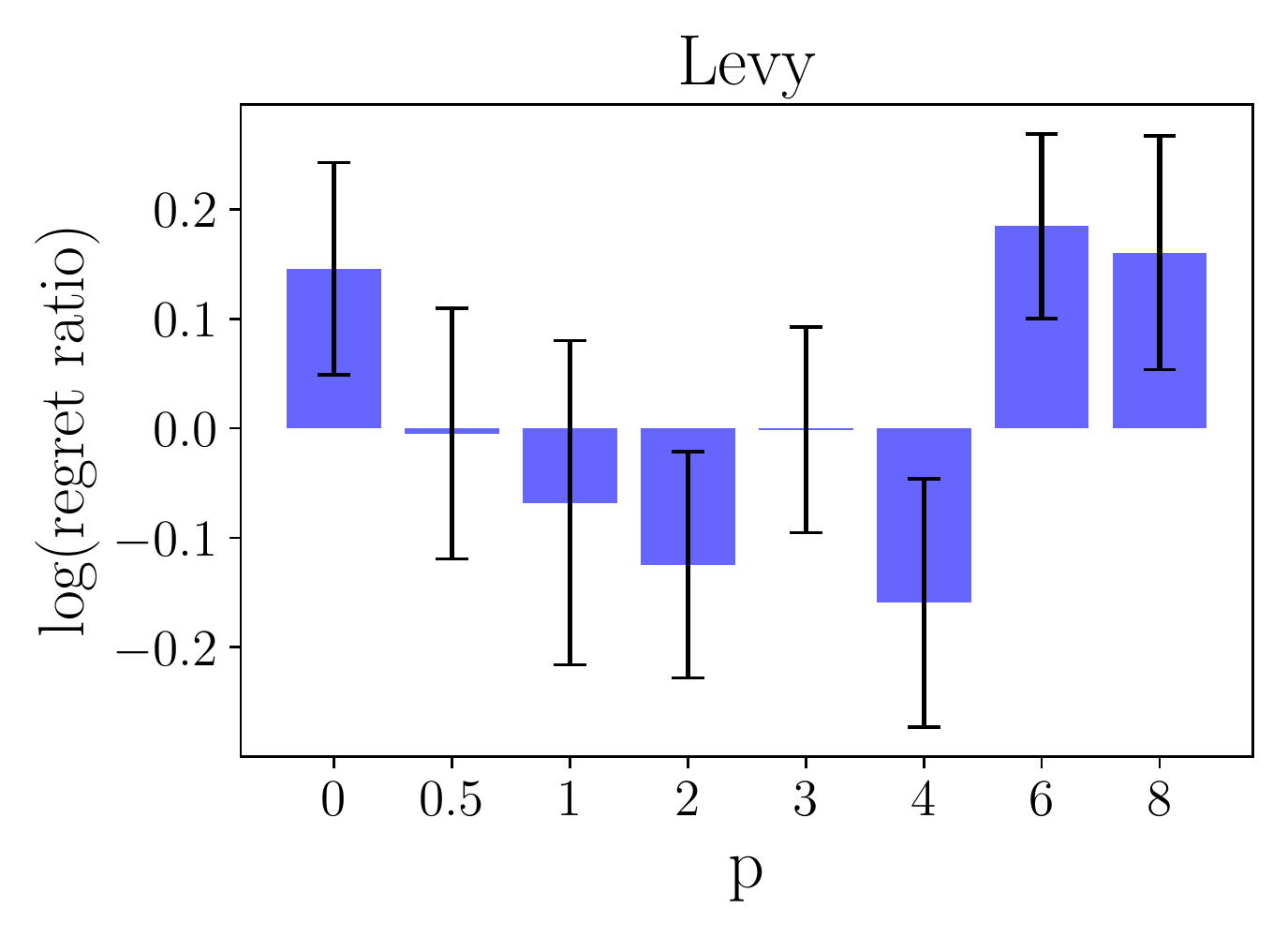}
	\includegraphics[height=.23\textheight]{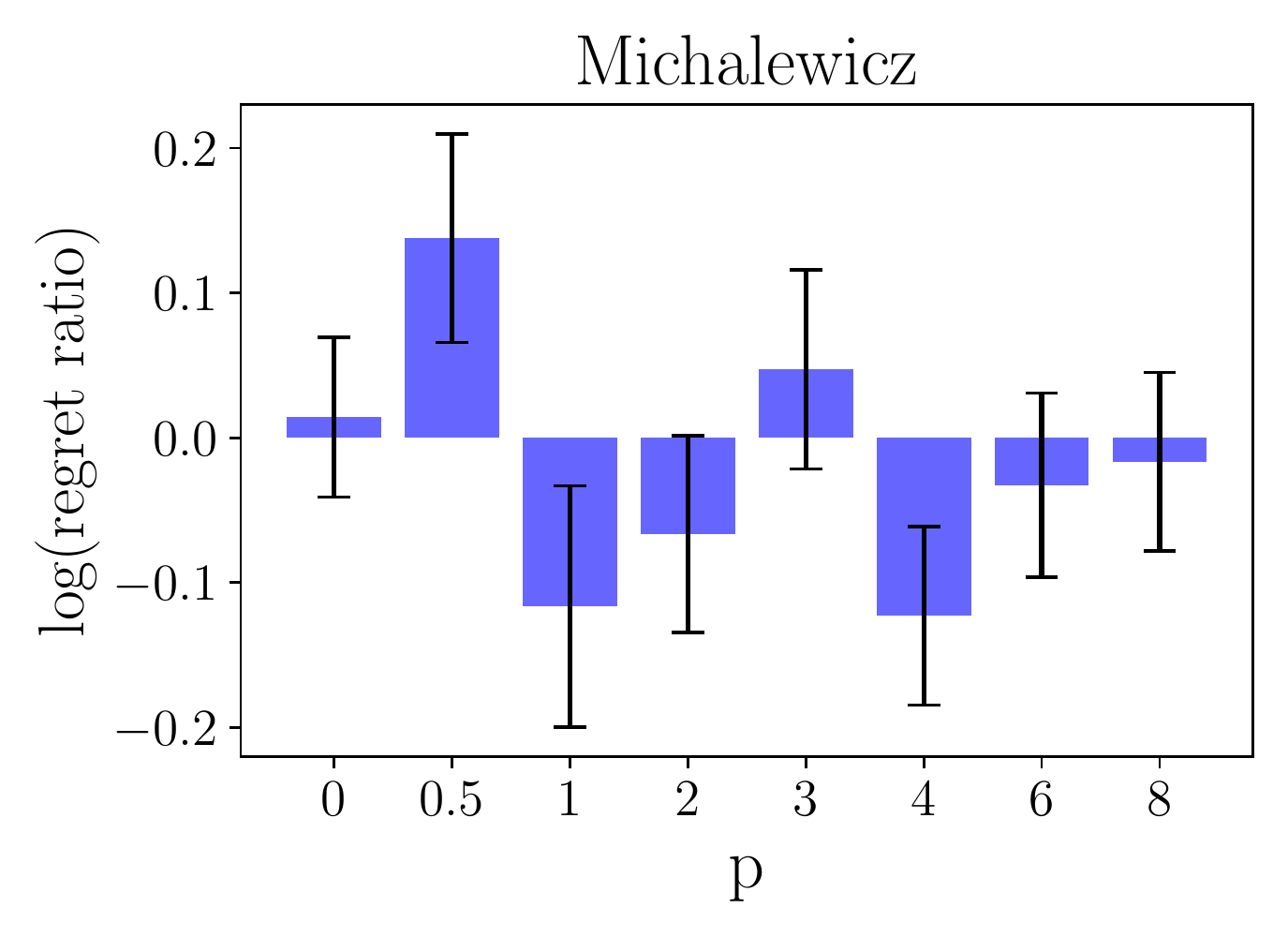}
	\caption{\label{fg:tpvsgp}Comparison of TP and GP for BO using $\alpha_p$ as the acquisition function. The vertical axis is the index \eqref{eq:fdsfp}, the negative value of which indicates that TP outperforms GP. The error bars represent one standard deviation of the mean.}
\end{figure}

Figure~\ref{fg:tpvsgp} summarizes the result of our numerical experiments. For the Himmelblau function, TP substantially outperforms GP. On the contrary, GP is better than TP for the Hartmann function (except for $p=0.5$). For the remaining four functions, we do not see any systematic trend on the relative performance of GP and TP. It is concluded that the use of TP is not always a better choice than GP, the choice is strongly problem dependent, and it is sensitive to the value of $p$. It could also depend on the choice of the kernel function, the prior placed on $\nu$, and the noise level. It seems that a more thorough analysis is needed to draw a definitive conclusion about the utility of TP for BO.

\section{Conclusions}\label{sc:conc}
In this work, we put forward a new acquisition function, or infill criterion, for BO with Gaussian processes, including classical EI and PI as special cases. The benchmark tests on miscellaneous functions revealed that the proposed method is on par with,  and in some cases substantially outperforms EI. We also derived the generalization of  the proposed method to BO with Student-t processes. As a future direction of research we wish to explore an optimal scheme for calibrating the parameter $p$ inherent to our method.

\bibliography{draft_EI_v2.bbl}
\end{document}